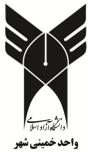

# آشکار سازی چشم در تصاویر دیجیتالی: چالش‌ها و راه‌حل‌ها

| میترا منتظری[1] | مهدیه منتظری | سعید سریزدی |

دانشجوی کارشناسی ارشد - دانشگاه شهید باهنر کرمان
mmontazeri@eng.uk.ac.ir

کارشناسی ارشد - دانشگاه امیرکبیر
montazeri@kmu.ac.ir

دانشیار - دانشگاه شهید باهنر کرمان
saryazdi@mail.uk.ac.ir

**چکیده**

آشکارسازی چشم در حوزه تشخیص هویت بیومتریک کاربرد مهمی دارد و یکی از روش‌های بازشناسی فرد شناخته شده است. در سال‌های اخیر تلاش‌های زیادی برای آشکارسازی چشم انجام شده است که بتوانند این آشکارسازی را به طور اتوماتیک و با شرایط مختلف تصویر انجام دهند. اما هر روش مشکلات و چالش‌های خاص خود را دارد و می‌تواند تعدادی از این شرایط را کنترل کند. در این مقاله انواع روش‌های مختلف برای آشکارسازی چشم به طور جامع دسته بندی و توضیح می‌گردد. در هر دسته بندی مزایا و معایب هر روش توضیح داده می‌شود.

**واژه‌های کلیدی:** بیومتریک، آشکارسازی چشم، عنبیه.

---

[1] عضو انجمن پژوهشگران جوان، دانشگاه شهید باهنر کرمان.

## 1- مقدمه

یکی از مهم ترین ویژگی‌های صورت چشم می باشد زیرا چشم نقش مهمی را برای داشتن سیستمی که بتواند به طور اتوماتیک صورت را تحلیل و بازشناسی کند، ایفا می کند. با پیدا کردن چشم بقیه ویژگی‌های صورت به آسانی آشکار می گردند[1]. مهم ترین مرحله آشکارسازی چشم، پیداکردن مدلی از چشم است که بتواند حوزه وسیعی از تصاویر صورت را کنترل کند. اما به دلیل داشتن شرایط مختلف تصویری زیاد، پیدا کردن مدل دقیقی که بتواند کارایی مناسبی داشته باشد تقریبا غیرممکن است. در نتیجه در سالهای اخیر روش‌های گوناگونی ارائه شدند که سعی در ایجاد روش‌هایی که از مدل خاصی پیروی نکنند و از ویژگیهای منحصر به فرد چشم استفاده کنند. از طرف دیگر، مشکلاتی نیز در تصاویر صورت وجود دارد که می توان آنها را به دو دسته کلی تقسیم کرد: دسته اول شامل مشکلاتی که در اثر شرایط محیطی ایجاد می شوند به عنوان مثال شرایط نوری، که باعث تغییر کیفیت تصویر یا روشن و تیره شدن آن می گردد یا نحوه قرار گرفتن دوربین، که باعث تغییر زاویه و اندازه تصویر گشته و منجر به تغییر در مدل چشم می شود. دسته دوم مربوط به شرایط خود فرد می باشد به عنوان مثال درجه باز بودن چشم در هنگام عکس برداری، نوع فیگور صورت شخص مانند غمگین یا خوشحال بودن، که باعث تأثیر بر روی چشم و مدل چشم می گردد و چرخش سر که این حالت می تواند در دسته اول نیز قرارگیرد.

از سال‌های گذشته روش‌های گوناگونی برای حل این مشکلات ارائه شده است. اما هر روش قادر است تعدادی از این مشکلات را حل کند. به طور کلی روش های آشکاری سازی چشم به سه دسته تقسیم می شوند [2]: دسته اول روش‌های مبتنی بر شکل که خود به دو دسته مدل ساده و پیچیده تقسیم می گردند [3, 4]. دسته دوم که روش-های مبتنی بر ویژگی‌های چشم می باشند که به دو دسته سطح خاکستری و اعمال فیلتر تقسیم می شوند [5, 6]. دسته آخر روش های مبتنی بر ظاهر، که به دو دسته دامنه سطح خاکستری و اعمال فیلتر تقسیم می گردند[7, 8]. اما در سال های اخیر بیش تر روش‌ها به صورت ترکیبی از چند روش بالا مانند ویژگی چشم و شکل چشم ارائه شده اند که در بخش 5 چند روش آن را توضیح خواهیم داد. در بخش 2، 3 و 4 به ترتیب دسته اول و دوم و سوم را شرح خواهیم داد و در انتها در بخش 6 به نتیجه گیری و جمع بندی می پردازیم.

## 2- روش های مبتنی بر شکل

روش‌های مبتنی بر شکل از دو مولفه تشکیل شده اند: مدل هندسی چشم و معیار شباهت. در مولفه اول یک مدل هندسی برای چشم طراحی می شود که دارای پارامترهای اندازه گیری برای توصیف این مدل است. بسته به تعداد و دقت این پارامترها مدل چشم می تواند ساده یا پیچیده باشد. مولفه دوم یا معیار شباهت، نوع معیار برای تشابه بین مدل ایجاد شده و مدل استخراج شده را تعیین می کند که بتواند محل چشم را در صورت پیدا کند. مهمترین ویژگی این روش قابلیت کنترل شکل، اندازه و چرخش سر است.

### 1-2 مدل ساده چشم

مدل ساده چشم از پارامترهای کمی برای توصیف مدل چشم استفاده می‌کند و شامل دو مرحله است: مرحله اول نمره دهی است در این مرحله به همه ویژگی های استخراج شده یک نمره داده می‌شود. درنتیجه یکسری از ویژگی های خوب از سایر ویژگی‌ها تفکیک می گردد. مرحله دوم تطبیق کردن مدل است که ویژگی های انتخاب شده در مرحله قبل را با مدل مورد نظر تطبیق می دهند. مشکل این روش‌ها این است که نمی توانند تغییرات برخی ویژگی‌های چشم مثل گوشه های چشم، پلک و یا حتی ابرو را مدل کنند و نیازمند تصویر با کنتراست بالا می باشند و حتی گاهی برای استخراج برخی ویژگی ها نیازمند آستانه مناسب هستند.

### 2-2- مدل پیچیده چشم

مدل پیچیده چشم از پارامترهای زیادتری نسبت به مدل قبل به منظور توصیف دقیق تر چشم استفاده می کند. مدل معروفی که در این زمینه ارائه شده مدل Yuille [9] است که از 11 پارامتر برای توصیف مدل چشم استفاده می نماید. این مدل شامل دو سهمی برای نمایش پلک ها و یک دایره برای نمایش عنبیه است. سپس این مدل با استفاده از یک سری قوانین، به تصویر اعمال می‌گردد به گونه‌ای که تابع انرژی برای دره‌ها، لبه‌ها، قله‌ها و غیره حداقل شود. شکل 1 این مدل و مراحل تطبیق این مدل با چشم را نشان می‌دهد.

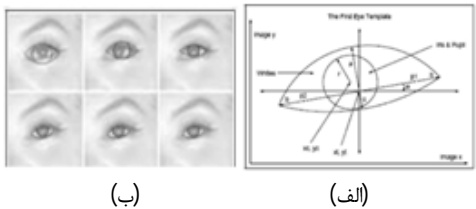

شکل (1) آشکارسازی چشم براساس مدل پیچیده چشم. (الف)مدل Yuille. (ب) آشکارسازی چشم با اعمال این مدل.

به طور کلی روش‌های مبتنی بر شکل دارای مشکلاتی از این قبیل هستند:

- محاسبات زیادی دارند.
- نیازمند تصویر با کنتراست بالا هستند.
- نیازمند مکان اولیه چشم که بایستی در نزدیکی چشم باشد، هستند.
- این روش نمی تواند به خوبی، چشم با فیگور های مختلف یا بسته بودن آن را پیدا کند.

به دلیل این که این روش نمی تواند حوزه وسیعی از تغییرات مدل چشم را کنترل کند از روش های مبتنی بر ویژگی استفاده می‌شود.

## 3- روش های مبتنی بر ویژگی

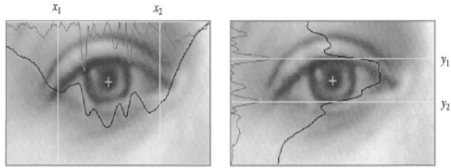

شکل (۲): (الف)مدل چشم. (ب)آشکار سازی مرکز عنبیه با یافتن نقاط مهم توسط تابع افکنش [۱۱]

### ۳-۲- ویژگی های چشم مبتنی بر اعمال فیلتر

این روش‌ها با اعمال فیلترهای خاص سعی دارند ویژگی‌های چشم را برجسته کنند. به عنوان مثال در مرجع [۱۲] از یک ماسک (شکل ۳) با دو فیلتر برای مردمک و اطراف آن استفاده شده است. در این روش از فیلتر ترتیبی با آلفای ۰/۶ برای مردمک و ۰/۴ برای اطراف آن استفاده کرده است. هدف از به کار نبردن فیلتر میانگین (آلفای ۰/۵)، داشتن اطمینان کافی از عدم وجود جواب‌های غلط مانند اشیاء گردی که اندازه ای کوچکتر از مردمک دارند، بوده است. اندازه ماسک با توجه به اندازه های استاندارد تعیین شده بود. پس از اعمال فیلتر تعدادی اشیاء پیدا می گردند. (شکل ۴-الف).

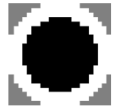

شکل (۳) فیلتر استفاده شده برای آشکارسازی مردمک و اطراف آن در مرجع [۱۲]

سپس با توجه به این که تصویر ممکن است حاوی نویز باشد از یک آستانه برای افزایش دامنه اشیاء پیدا شده استفاده می کند. همان طورکه در شکل ۴-ب مشاهده می شود، منجر به آشکارشدن اشیاء بیشتری می‌گردد. در نهایت اشیاء نادرست را با اعمال یکسری محدودیت ها به عنوان مثال فاصله بین دو چشم حذف می کند.(۴-ج و ۴-د).

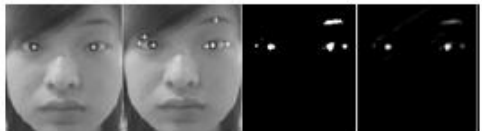

شکل (۴): مراحل پیدا کردن چشم در مرجع [۱۲]

منتظری و همکاران در مرجع [۱۳]، روشی جدید ارائه دادند که از هر دو روش مبتنی بر ویژگی بر سطح خاکستری و اعمال فیلتر استفاده می‌کند. در این روش در ابتدا برای آشکار سازی صورت از ترکیب افکنش واریانس و افکنش انتگرال استفاده شده است. البته قبل از انجام این مرحله با استفاده از عملگر مورفولوژی تصویر ورودی بهبود می یابد. سپس پنجره چشم با استفاده از نگاشت عمودی مشخص

روش‌های مبتنی بر ویژگی با بررسی مشخصات چشم سعی در پیدا کردن ویژگی های برجسته چشم مانند انعکاس خاص قرنیه نسبت به نور یا تغییر شدید روشنایی به تاریکی از ناحیه اطراف چشم به عنبیه را دارند. این روش‌ها سعی دارند از ویژگی های محلی چشم و صورت که به تغییرات نور حساس نباشند استفاده کنند [۲].

### ۳-۱- ویژگی محلی چشم مبتنی بر سطح خاکستری

این روش‌ها از ویژگی هایی مبتنی بر سطح خاکستری استفاده می-کنند. بسیاری از این روش ها از ویژگی تغییر شدید سطح خاکستری در چشم استفاده می کنند. در مرجع [۱۰] از تابع افکنش واریانس برای پیدا کردن این ویژگی و در نتیجه آشکارسازی عنبیه چشم استفاده نموده است در این روش از فرمول (۱) و (۲) در راستای افق و عمود استفاده کرده است.

$$V_\mathrm{m}(x) = \frac{1}{y_2 - y_1}\int_{y_1}^{y_2} I(x,y)\, \mathrm{d}y, \qquad (۱)$$

$$H_\mathrm{m}(y) = \frac{1}{x_2 - x_1}\int_{x_1}^{x_2} I(x,y)\, \mathrm{d}x. \qquad (۲)$$

افکنش انتگرال به دلیل حساس نبودن به تغییرات نمی تواند برای آشکارسازی کافی باشد. در نتیجه از افکنش واریانس که نسبت به تغییرات حساس است استفاده نموده است. در مرجع [۱۱]، از ترکیب افکنش واریانس و افکنش انتگرال برای آشکارسازی عنبیه استفاده شده است در این مقاله نشان داده شده است که افکنش انتگرال می تواند بر افکنش واریانس برتری داشته باشد، چون واریانس صرفاً تغییرات را نشان می دهد. در نتیجه بر این ایده تکیه دارد که اگر این دو تابع با یکدیگر ترکیب شوند می‌توانند تأثیر بهتری در آشکارسازی داشته باشند که این تأثیر در انواع پایگاه داده ها نشان داده شده است. در شکل ۲-الف مدل چشم استفاده شده در این مقاله را نشان می دهد و در شکل ۲-ب نحوه پیدا کردن نقاط مهم در پیدا کردن عنبیه را نشان می دهد. خطوط تیره تابع افکنش و خطوط خاکستری مشتق اول این تابع را نشان می دهد. ثابت شده است این روش نسبت به چرخش و اندازه مقاوم است و به دلیل وجود تابع افکنش واریانس می تواند نسبت به سایه‌ها مقاوم باشد. ولی این روش در حالتی که قسمتی از چشم در اثر مو یا چرخش نامشخص باشد باشکست روبه رو می شود.

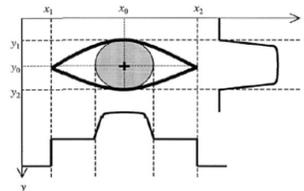

(الف)

می‌گردد. در نهایت از دو ماسک برای آشکارسازی مرکز مردمک استفاده شده است. دقت 99.5 این روش حاکی از این مطلب بود که این روش می‌توانست شرایط مختلف تصویر را به خوبی کنترل کند.

در حالت کلی روش‌های مبتنی بر ویژگی نسبت به برخی ویژگی‌ها (ویژگی‌هایی که با تغییر شرایط مقاوم نیستند مانند لبه یا بسته بودن چشم) دچار مشکل می‌شود. در نتیجه روش‌های دیگری که مبتنی بر ظاهر چشم می‌باشند استفاده می‌شود.

### 4- روش‌های مبتنی بر ظاهر

روش‌های مبتنی بر ظاهر از تعدادی تصویر با اعمال فیلتر یا تغییر سطح خاکستری برای تولید تصویر نمونه استفاده می‌کنند. این روش‌ها از تعداد زیادی نمونه آموزشی که شرایط نوری یا چرخش متفاوت دارند، استفاده می‌کنند.

### 1-4 دامنه سطح خاکستری

در این روش از اطلاعات سطح خاکستری برای تولید تصویر نمونه استفاده می‌گردد. از این روش در مرجع [14] استفاده شده است. در این روش از الگوریتم ژنتیک برای آشکارسازی چشم استفاده گردیده است. تصویر نمونه (شکل 5-الف) با میانگین گرفتن 20 تصویر چشم تولید شده است و از دو پارامتر مقیاس و چرخش برای آشکارسازی بیش‌تر استفاده نموده است.(شکل 5-ب)

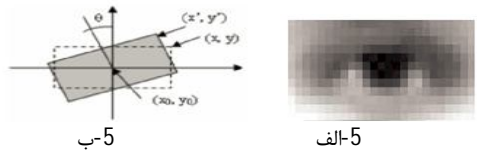

شکل 5(الف)تصویر نمونه،(ب) دو پارامتر مقیاس و چرخش [14]

ساختار کروموزوم از 4 قسمت شامل دو قسمت اول موقعیت مرکز چشم و دو قسمت بعدی مقیاس و چرخش آن تشکیل شده است. تابع شایستگی همبستگی بین نمونه داده شده و ناحیه پیدا شده در تصویر دارد. پس از اجرای کامل الگوریتم ژنتیک، جواب بهینه، موقعیت چشم در تصویر را مشخص می‌کند.

### 2-4 اعمال فیلتر

برخی از روش‌ها از اعمال فیلتر برای تولید تصویر نمونه استفاده می‌کنند.[15] اعمال فیلتر به کار برده شده در این روش با روش فیلتر در روش مبتنی بر ویژگی متفاوت است. در روش‌های مبتنی بر ویژگی با اعمال فیلتر سعی در برجسته کردن ویژگی‌های خاص می‌باشد ولی در این روش تصویر اعمال شده توسط فیلتر را به عنوان تصویر نمونه آشکارسازی استفاده می‌گردد.

روش‌های مبتنی بر ظاهر دارای مشکلاتی از این قبیل هستند: ذاتاً با تغییر مقیاس و چرخش مشکل دارند، با چرخش صورت دچار مشکل می‌شود، به دلیل این که مدل مستقیمی از چشم ندارد، بنابراین پارامتر خاصی از چشم ندارد و درنهایت نمی‌تواند رنجی از تغییرات را کنترل نماید. برای حل آن فقط باید از مدل صورت ثابت یا از چند تصویر نمونه استفاده شود.

### 5- روش‌های ترکیبی

در سال‌های اخیر بیش‌تر روش‌ها به صورت ترکیبی از چند روش بالا مانند ویژگی و شکل چشم ارائه شده‌اند. در مرجع [16] شکل و ویژگی برای آشکارسازی استفاده گردیده است. این روش شامل سه مرحله می‌باشد. مرحله اول شامل آشکارسازی مرکز مردمک است و از واکنش فیزیکی خاص مردمک چشم نسبت به نور استفاده نموده است. این واکنش در محیط HSV در مولفه‌های H به خوبی مشخص می‌باشد که به صورت یک لکه روشن نشان داده می‌شود. سپس با استفاده از افکنش واریانس موقعیت این لکه روشن و در نتیجه مرکز مردمک مشخص می‌گردد.(شکل 6)

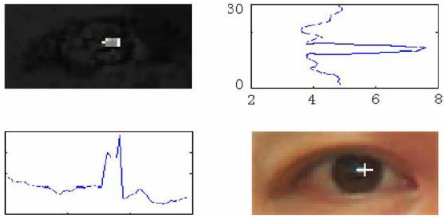

شکل (6): آشکارسازی مرکز مردمک [16]

درمرحلۀ دوم لبه خارجی عنبیه آشکار می‌گردد. برای آشکارسازی عنبیه ابتدا یک دایره به مرکز بدست آمده در مرحله قبل ایجاد می‌شود، سپس این دایره تا جایی که میانگین سطح خاکستری حداقل شود جابه‌جا می‌گردد (زیرا مردمک جسمی تیره است). با کوچک و بزرگ کردن لبه‌های دایره سعی می‌شود که میانگین سطح خاکستری کمتر گردد.(شکل 7).

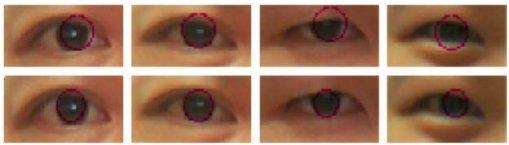

شکل(7): آشکارسازی لبه خارجی عنبیه [16]

مرحله آخر، آشکارسازی پلک‌های چشم است که این کار ابتدا با آشکارسازی دو گوشه چشم با استفاده از اعمال فیلتر گابور (شکل8) انجام و سپس با آشکارسازی دو نقطه وسط پلک‌ها که روشنی بالاتر دارند منحنی پلک چشم بدست می‌آید(شکل9).

## ۶- نتیجه‌گیری و جمع بندی

در این مقاله روش‌های مختلف آشکارسازی چشم معرفی و دسته بندی گردید. مزایا و معایب هر روش معرفی شد. در هر دسته بندی چند مرجع که از روش مربوطه استفاده می‌نمودند، معرفی گردید. در پایان به دو نکته مهم می‌توان اشاره کرد:

- هر روش برای آشکارسازی چشم مزایا و معایب و درنتیجه کاربرد خاص خود را دارد، درنتیجه بسته به نوع تصاویر بایستی روش مناسب را به کار بست.
- برای یافتن روشی که بتواند دامنه وسیعی از شرایط را کنترل کند بایستی از روش‌های ترکیبی استفاده نمود.

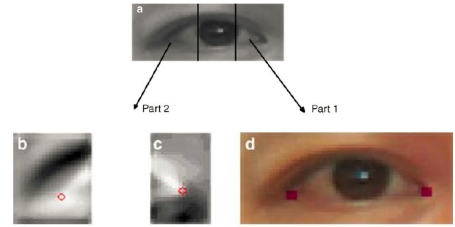

شکل(۸): آشکارسازی دو گوشه چشم [۱۶]

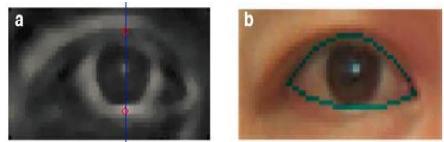

شکل (۹): مراحل آشکارسازی پلک‌های چشم [۱۶]

در مرجع [۱۷]، برای آشکارسازی چشم ابتدا محدوده‌ی صورت را با استفاده از لبه های افقی و عمودی بدست می آورد. سپس برای آشکارسازی چشم از این نکته استفاده می شود که چشم دارای لبه های عمودی بیش تری نسبت به ابرو می باشد و یک پنجره با ارتفاع ۱۵± در اطراف خط احتمالی از وسط چشم(EC) ایجاد می کند. چون این امکان وجود دارد که EC به طور دقیق از وسط چشم عبور نکرده باشد، لبه های افقی ناحیه بالای خط EC و پایین خط EC محاسبه می گردد و خط EC به طرف تراکم بالاتر انتقال داده می شود. در نهایت با استفاده از یک شبکه عصبی به گونه ای که در ابتدا شبکه را با نمونه‌های + و – آموزش می دهد. نمونه + نمونه ای است که درآن چشم کاملاً در پنجره قرار دارد و نمونه منفی نمونه ای است که درآن قسمتی از چشم در پنجره وجود دارد. سپس با حرکت پنجره بر روی تصویر چشم و بااستفاده از معیارهای eigenvalues, eigenvectors به عنوان ورودی شبکه عصبی، محل دقیق تر را پیدا می کند. در مرجع [۱۸] برای آشکارسازی چشم ابتدا تصویر را قسمت بندی نموده که این قسمت بندی تا جایی است که به ناحیه همگن برسیم . چون چشم واریانس بالایی دارد در این ناحیه قسمت بندی زیادی داریم( شکل۱۰). سپس این بلاک‌ها را بااستفاده از آستانه ویژه ادغام می‌نماییم. این آستانه که نقش مهمی را ایفا می‌نماید، به صورت تجربی از مقادیر ومحیط های متفاوت با اختلاف بین عنبیه و سفیدی اطراف آن بدست می آید. بااستفاده از عملگر مورفولوژی نقاط نزدیک را ادغام نموده ودرنهایت باتوجه به این که چشم درنیمه بالایی صورت است بزرگترین ناحیه را به عنوان چشم تشخیص می‌دهد.

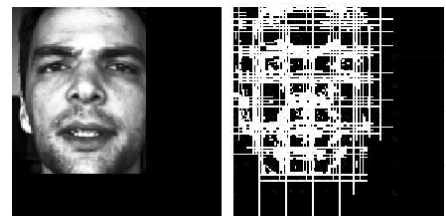

شکل۱۰ قسمت بندی تصویر [۱۸]